\documentclass{article}

\usepackage{microtype}
\usepackage{graphicx}
\usepackage{subfigure}
\usepackage{booktabs} 
\usepackage{xcolor}
\usepackage{color}
\usepackage{caption}
\usepackage{amsmath, amsthm, amssymb, bm, bbm}

\usepackage{amsmath,amsfonts,bm}

\newcommand{\bomega}{{\boldsymbol{\omega}}}

\def\secref#1{section~\ref{#1}}

\def\eqref#1{equation~\ref{#1}}

\def\1{\bm{1}}

\DeclareMathAlphabet{\mathsfit}{\encodingdefault}{\sfdefault}{m}{sl}
\SetMathAlphabet{\mathsfit}{bold}{\encodingdefault}{\sfdefault}{bx}{n}

\def\sR{{\mathbb{R}}}

\newcommand{\bx }{\boldsymbol{x}}
\newcommand{\bX}{\boldsymbol{X}}

\newcommand{\bff}{\boldsymbol{f}}
\newcommand{\bk}{\boldsymbol{k}}
\newcommand{\bK}{\boldsymbol{K}}

\newcommand{\by}{\boldsymbol{y}}

\usepackage{multirow}

\usepackage{hyperref}
\usepackage{url}            

\usepackage[accepted]{icml2021}

\icmltitlerunning{GP-Tree: A Gaussian Process Classifier for Few-Shot Incremental Learning}

\newcommand{\ignore}[1]{}

\newcommand{\figrref}[1]{Figure~\ref{#1}}
\newcommand{\tblref}[1]{Table~\ref{#1}}
\newcommand{\eqnref}[1]{Eq.~\ref{#1}}
\newcommand{\secnref}[1]{Section~\ref{#1}}

\newcommand{\PG}{P\'olya-Gamma }
\newcommand{\pg}{P\'olya-Gamma }
\newcommand{\normal}{\mathcal{N}}
\newcommand{\bld}[1]{\boldsymbol{#1}}
\newcommand{\e}{\mathbb{E}}

\newcommand{\Kmm}{\bld{K}_{mm}}
\newcommand{\Kmn}{\bld{K}_{mn}}
\newcommand{\Kmi}{\bld{K}_{mi}}
\newcommand{\Knm}{\bld{K}_{nm}}
\newcommand{\Kim}{\bld{K}_{im}}
\newcommand{\Qnn}{\bld{Q}_{nn}}
\newcommand{\KmmInv}{\bld{K}_{mm}^{-1}}
\newcommand{\fbar}{\bar{\bld{f}}}
\newcommand{\BigOmega}{\bld{\Omega}}
\newcommand{\BigLambda}{\bld{\Lambda}}
\newcommand{\BigSigma}{\bld{\tilde{\Sigma}}}

\begin{document}

\twocolumn[
\icmltitle{GP-Tree: A Gaussian Process Classifier for Few-Shot Incremental Learning}




\begin{icmlauthorlist}
\icmlauthor{Idan Achituve}{biu}
\icmlauthor{Aviv Navon}{biu}
\icmlauthor{Yochai Yemini}{biu}
\icmlauthor{Gal Chechik}{biu,nvidia}
\icmlauthor{Ethan Fetaya}{biu}
\end{icmlauthorlist}

\icmlaffiliation{biu}{Bar-Ilan University, Israel}
\icmlaffiliation{nvidia}{Nvidia, Israel}

\icmlcorrespondingauthor{Idan Achituve}{Idan.Achituve@biu.ac.il}

\icmlkeywords{Gaussian Process Classification, Incremental Learning, Deep Kernel Learning}

\vskip 0.3in
]



\printAffiliationsAndNotice{}  

\begin{abstract}
    Gaussian processes (GPs) are non-parametric, flexible, models that work well in many tasks. Combining GPs with deep learning methods via deep kernel learning (DKL) is especially compelling due to the strong representational power induced by the network. However, inference in GPs, whether with or without DKL, can be computationally challenging on large datasets. Here, we propose \textit{GP-Tree}, a novel method for multi-class classification with Gaussian processes and DKL. We develop a tree-based hierarchical model in which each internal node of the tree fits a GP to the data using the P\`olya-Gamma augmentation scheme. As a result, our method scales well with both the number of classes and data size. We demonstrate the effectiveness of our method against other Gaussian process training baselines, and we show how our \textit{general} GP approach achieves improved accuracy on standard incremental few-shot learning benchmarks.
\end{abstract}

\section{Introduction}
\label{sec:intro}
Gaussian processes (GPs) are a popular Bayesian non-parametric approach that enjoys a closed-form marginal likelihood, thus avoiding one of the major computational difficulties of many Bayesian approaches. However, due to several obstacles, successfully applying GPs to certain problems may be challenging. First, the performance of GP models heavily depends on the kernel function being used. In domains such as images, where common kernels are not a good measure of semantic similarity, this can hinder the performance. This problem is commonly addressed by deep kernel learning (DKL) where GPs are combined with modern neural networks to learn a useful kernel function from the data \cite{calandra2016manifold, gordon16_DKL}.

Second, computing the marginal likelihood involves storing and inverting an $n\times n$ kernel matrix, where $n$ is the number of training examples. This can limit exact inference on large datasets, especially when combined with DKL where the kernel needs to be re-computed at each iteration. A common approach to handle large datasets is to learn a small set of inducing points, which act as a proxy for the training data \cite{silverman1985some,quinonero2005unifying, sneldon_Gharamani_IP}.

While extensive work has been done on handling these challenges for Gaussian process regression, comparatively little has been done on how to scale Gaussian process classification \citep[see][for a recent review]{liu2020gaussian}. This is partly because the categorical distribution on the target variable results in non-Gaussian posteriors and we no longer have a closed-form marginal likelihood. One appealing
approach to address this obstacle is the \pg augmentation \cite{polya_gamma}. In the \pg augmentation, the posterior becomes Gaussian when conditioned on the augmented \pg variable. However, this augmentation was designed for binary classification tasks. Since then, several extensions to multi-class classification have been proposed \cite{linderman2015dependent, galy2020multi, snell2020bayesian}. However, as we will show empirically their performance degrades as the number of target classes increases.

In this study, we present a novel method for Gaussian process classification (GPC) that is designed to handle both small and large datasets. Importantly, it is also designed to handle a large number of classes. We develop a tree-based model in which each node solves a binary classification task using a Gaussian process and the \PG augmentation scheme. We term our method \textit{GP-Tree}. GP-Tree has great flexibility as it allows us to take samples from the posterior with Gibbs sampling, or apply variational inference and use the inducing points approximation. To train GP-Tree on large-scale image classification tasks, we further combine it with DKL and we show how in this setup as well it is superior to popular GPC methods.

Finally, we apply GP-Tree to incremental few-shot learning challenges \cite{tao2020fscil}. In incremental few-shot learning, we assume that the data come sequentially. Initially, the learner is presented with many samples from some base classes. Then, at each new iteration, it has access only to a new, small, dataset that originated from a set of novel classes not seen during previous iterations. Here the challenges are two-fold, generalizing from a small number of training points for the novel classes and avoiding catastrophic forgetting of the previously encountered classes. We claim that GPs are a natural fit for this problem. The inducing points, which summarize the training data, can help mitigate the catastrophic forgetting problem, and GPs generalize well from small datasets due to their Bayesian nature. Indeed, we show that once we get GPs to scale and successfully train on the base classes, our GP approach achieves performance gains over baseline methods on incremental few-shot benchmarks.


Thus, we make the following novel contributions: (1) we show how current Gaussian process classification methods struggle when the number of classes to be learned is large; (2) we present a novel method for Gaussian process classification that is designed to handle a large number of classes and large datasets based on the \pg augmentation; (3) we present GPs as a promising new research direction for few-shot incremental learning; (4) we demonstrate competitive classification accuracy on two benchmark datasets designed for few-shot incremental learning. Our code is publicly available at \url{https://github.com/IdanAchituve/GP-Tree}.  



\section{Background} \label{sec:background}
\subsection{Notations}
We denote vectors with bold lower-case font, e.g. $\bx$, and matrices with capital bold font, e.g. $\bX$. Given a dataset $(\bx_1,y_1),...,(\bx_n,y_n)$, we denote by $\by=[y_1,...,y_n]^T$ the vector of labels, and by $\bX\in\sR^{n\times d}$ the design matrix whose $i^{th}$ row is $\bx_i$. In the classification case, each $y_i$ takes a value from $\{1,...,C\}$ class labels.
\subsection{Gaussian Processes}
In Gaussian process learning we assume the mapping from the input points to the target values is via a latent function $f$. The target values are assumed to be independent when conditioned on the latent function, i.e., $p(\by|\bX,f)=\prod_{i=1}^np(y_i|f(\bx_i))$.
The latent function is assumed to follow a Gaussian process prior  $f\sim\mathcal{GP}(m(\bx),~k(\bx,\bx'))$, where the evaluation vector of $f$ on $\bX$,  $\bff=[f(\bx_1),...,f(\bx_n)]^T$, has a Gaussian distribution $\bff\sim\mathcal{N}(\boldsymbol{\mu},~\bK)$, where $\boldsymbol{\mu}_i=m(\bx_i)$ and $\bK_{ij}=k(\bx_i,\bx_j)$. The mean $m(\bx)$ is commonly taken to be the constant zero function, and the kernel $k(\bx,\bx')$ is a positive semi-definite function. 

Let $\bX,\by$ be the training data, and let $f_*$ be the evaluation of $f$ on a novel point $\bx_*$. In the regression case, we assume $p(y|\bx,f)=\normal(f(\bx),~\sigma^2)$. Therefore, the predictive distributions, $p(f_*|\bx_*, \by, \bX)$ and
$p(y_*|\bx_*, \by, \bX)=\int p(f_*|\bx_*, \by, \bX)p(y_*|f_*)df_*$, are Gaussians with known parameters. Specifically,
\begin{equation}
    \begin{aligned} 
    &p(f_*|\bx_*, \by, \bX)=\mathcal{N}(\mu_*,~\sigma_*),\\
    &\mu_*=\bk_{*}^T(\bK+\sigma^2 \bld{I})^{-1}\by,\\
    &\sigma_* = k_{**} - \bk_*^T(\bK+\sigma^2\bld{I})^{-1}\bk_*.
    \end{aligned} 
\end{equation}
Where, $k_{**} = k(x_*, x_*)$, and $\bk_*[i]=k(\bx_i, \bx_*)$.
This closed-form solution allows us to avoid the costly marginalization step; however, it entails the inversion of an $n\times n$ matrix which can be expensive to compute for large datasets.

DKL \cite{gordon16_DKL} is a popular choice to apply a kernel on structured data such as images. The kernel over the input data points is commonly in the form of a fixed kernel on an embedding learned by a deep neural network $g_\theta$, e.g., $k_\theta(\bx,\bx')=\exp(-||g_\theta(\bx)-g_\theta(\bx')||^2/{2\ell^2})$. Therefore, the closed-form inference is of even greater importance as it allows to easily backpropagate through the GP inference. 
\subsection{The \pg Augmentation} \label{sec:backgroud_pg}
When applying GPs to classification tasks, the likelihood $p(y|f(\bx_i))$ is no longer Gaussian. The predictive distributions are also no longer Gaussian and we do not have a closed-form solution for them. To overcome this limitation, 
several methods were offered based on the \pg augmentation \cite{polya_gamma} to model the discrete likelihoods in GPs \cite{linderman2015dependent, WenzelGDKO19, galy2020multi, snell2020bayesian}. The \pg augmentation hinges on the following identity 
\begin{equation}\label{eq:polya_gamma}
    \frac{(e^{\psi})^a}{(1+e^{\psi})^b}=2^{-b}e^{\kappa\psi}\mathbb{E}_{\omega}[e^{-\omega\psi^2/2}],
\end{equation}
where $\kappa=a-b/2$, and $\omega$ has the \pg distribution $\omega\sim PG(b,0)$.

Suppose we have a binary classification task with $y \in \{0, 1\}$, and we are given a vector of latent function values $\bff$, the likelihood can be written as, 
\begin{equation}\label{eq:node_likelihood}
    \begin{aligned}
    p(\by | \bff) &= \prod_{j = 1}^{n} \sigma(f_j)^{y_j} (1 - \sigma(f_j))^{1 - y_j} = \prod_{j = 1}^{n} \frac{e^{y_jf_j}}{1 + e^{f_j}},
    \end{aligned}
\end{equation}
where $\sigma(\cdot)$ is the sigmoid function. We can now use \eqnref{eq:polya_gamma} to introduce auxiliary \pg variables (one per sample) such that we recover \eqnref{eq:node_likelihood} by marginalizing them out. If instead we sample the \pg variables $\bomega$ we get the following posteriors:
\begin{equation}\label{eq:posterior_dist}
    \begin{aligned}
        p(\bff | \by, \bomega) &= \normal(\bff | \bld{\Sigma}(\bld{K}^{-1}\bld{\mu} + \bld{\kappa}),~\bld{\Sigma}),\\
        p(\bomega | \by, \bff) &= PG(\bld{1},~\bff).
    \end{aligned}
\end{equation}
Where $\kappa_j = y_j - 1/2$, $\bld{\Sigma} = (\bld{K}^{-1} + \bld{\Omega})^{-1}$, and $\bld{\Omega} = diag(\bld{\omega})$. We can now sample from $p(\bff,\bomega|\by, \bX)$ using block Gibbs sampling and get Monte-Carlo estimations of the marginal and predictive distributions.

\subsection{Inducing Points}\label{sec:inducing_points}
Exact inference with GPs compels us to store the entire training set and invert an $n\times n$ matrix $\bK_{nn}$, which greatly limits their use. A common solution to this problem is to use inducing points \cite{silverman1985some, quinonero2005unifying, sneldon_Gharamani_IP}. With inducing points, we usually define $m \ll n$ pseudo-inputs whose locations are often trainable parameters. Then we only need to invert an $m\times m$ matrix, thus allowing us to control the computational cost.


Our approach closely follows the inducing points method for binary classification with the \pg augmentation presented in \cite{WenzelGDKO19}. We define $\bar{\bX}$ as the inducing locations/inputs and $\bar{\bff}$ as the latent function value evaluated on $\bar{\bX}$. We have,
\begin{align}
        &p(\bar{\bff})=\mathcal{N}(0,~\bK_{mm}),
\end{align}
\begin{equation}
    \begin{aligned} 
    &p(\bff|\bar{\bff})=\mathcal{N}(\bK_{nm}\bK_{mm}^{-1}\bar{\bff},~\boldsymbol{Q}_{nn}),\\
    &\boldsymbol{Q}_{nn}=\bK_{nn}-\bK_{nm}\bK_{mm}^{-1}\bK_{mn}.
    \end{aligned} 
\end{equation}
\begin{align}
    &p(\by,\bomega,\bff,\bar{\bff})=p(\by|\bff,\bomega)p(\bff|\bar{\bff})p(\bar{\bff})p(\bomega),
\end{align}
where $\bK_{mm}$ is the kernel matrix on the inducing locations, and $\bK_{nm}$ is the cross-kernel matrix between the inputs and the inducing locations. While $p(\bar{\bff}|\by,\bomega)$ is Gaussian due to the \pg augmentation, the distribution after marginalizing over $\bomega$, $p(\bar{\bff}|\by)$, is not Gaussian. Following \cite{hensman2015scalable},
\citet{WenzelGDKO19} proposed a variational inference approach based on the following assumption $q(\bomega,\bar{\bff})=q(\bomega)q(\bar{\bff})$, where $q(\bomega)$ is a \pg density and $q(\bar{\bff})$ is a Gaussian density. Then, the inducing locations and the variational parameters are learned with the evidence lower bound.

\section{Method} \label{sec:method}
\begin{figure*}[t]
\centering
    \begin{subfigure}[Multinomial Stick Break Tree]{
    \includegraphics[width=0.3\linewidth]{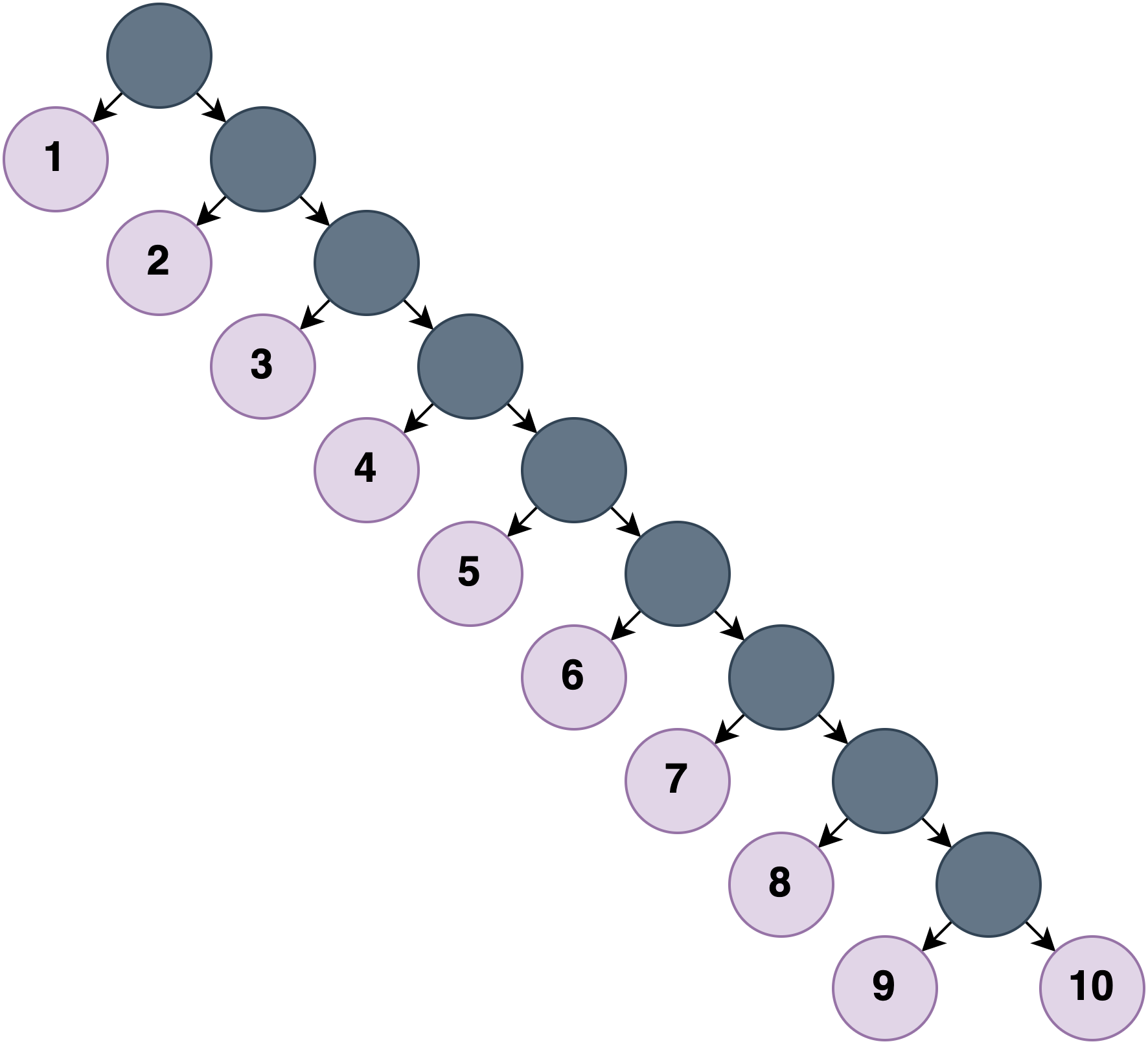}
    \label{fig:sb_tree}
    }
    \end{subfigure}
    \hfill
    \begin{subfigure}[GP-Tree]{
    \includegraphics[width=0.6\linewidth]{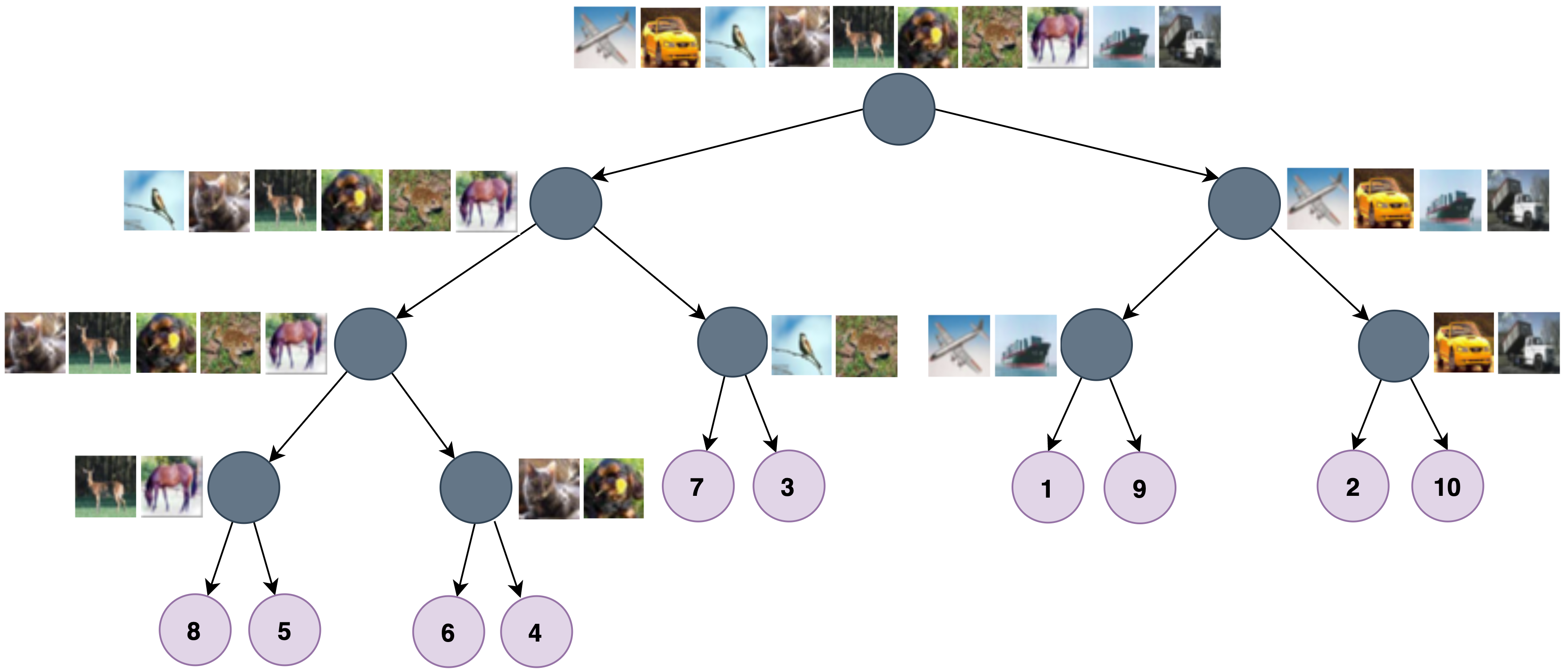}
    \label{fig:gp_tree}
    }
     \end{subfigure}
    \caption{The trees corresponding to the multinomial stick break model (left) and the GP-Tree model (right) for CIFAR-10. The multinomial stick break generates an unbalanced tree in which the order of the classes is arbitrary. GP-Tree, on the other hand, generates a more balanced tree that is divided by the semantic meaning of the classes. For example, motorized vehicles are on the right subtree of the root node while animals are on the left one. This semantic partition is pronounced at all tree levels.}
    \label{fig:trees}
\end{figure*}

In Sections \ref{sec:hier_cls}-\ref{sec:learning_full} we describe our method to train GP classifiers with DKL that scale to large training sets and a large number of classes. Afterward, in \secnref{sec:method_incremental_learning}, we will show how our model, with minor modifications, can be adjusted to the few-shot class-incremental learning setup.

\subsection{Hierarchical Classification}
\label{sec:hier_cls}
Consider the case presented in \cite{linderman2015dependent} aimed at modeling categorical and multinomial data. To derive a \PG augmentation \citet{linderman2015dependent} utilized the stick-breaking representation for the multinomial density. 
They turn the multi-class classification task into a sequence of $C - 1$ binary classification tasks where $y^j=1$ if the original label is $j$ and $y^j=0$ if the original label is larger than $j$. 
Then, $C - 1$ independent Gaussian processes can be learned, one for each of the binary classification tasks. Denoting by $\bff_i = (f_i^1, ..., f_i^{C-1})$ the vector of latent processes values at the $i^{th}$ example, the stick-break likelihood is: 
\begin{align} \label{eq:stick_break_likelihood}
    & p(y_i=c|\bff_i) = \sigma(f^c)\prod_{k < c} (1 - \sigma(f^k)),
\end{align}
where the remaining probability mass is assigned to the $C^{th}$ class. While the stick-breaking formulation allows to break the multi-class classification problem into a sequence of  binary classification tasks, as we will show in \secnref{sec:exp_gibbs}, its performance degrades with the number of classes. Intuitively, the stick-breaking process can be viewed as a hierarchical classification with an extremely unbalanced tree (see illustration in \figrref{fig:sb_tree}).
This sequential structure can be severely sub-optimal for two reasons: (1) the number of binary classification tasks needed to classify a data point grows linearly with the number of classes instead of logarithmic for a perfectly balanced tree; (2) not all label splits result in equally hard binary classification tasks, yet the stick-breaking process uses the default label ordering.

Therefore, we propose to use a tree-structured hierarchical classification instead of the sequential alternative. This construction allows finding a tree structure that results in easy-to-learn binary tasks.
Conceptually, we create a tree by splitting the data recursively by classes until we get to single class leaves.
More formally, starting at the root node, we partition the classes $\{1,...,C\}$ into two disjoint sets $C_l$ and $C_r$, 
such that $C_l \cup C_r = \{1, ..., C\}$. Let $D_l$ and $D_r$ denote the data points associated with the classes in $C_l$ and the classes in $C_r$ respectively. We assign $D_l$ to the left child of the root node and $D_r$ to the right one.  We recursively apply the same operation at each node until we are left with single class leaf nodes. Thus, a binary tree is formed (not necessarily a perfectly balanced one), see \figrref{fig:gp_tree} for an example. We then fit a binary GP classifier to each internal node of the tree that makes a binary decision to either go left or go right at that node. 

The model quality depends on how we partition the data, namely the tree construction processes described above. A naive approach would be to use a random balanced binary tree; however, this strategy does not take advantage of the semantic meaning of the classes. We propose the following procedure, we first compute a representative prototype for each class by taking the mean of all samples (or their representation obtained by a NN) belonging to the same class coordinate-wise. We then normalize the vectors to have unit length and apply divisive hierarchical clustering. We recursively split each node by using $k$-means++ clustering \cite{arthur2007k} with $k=2$ on the class prototype vectors, until we are left with single class leaves. We note that partitioning the data in this manner may be sub-optimal when working on the input space directly; however, when applied on features extracted by a NN it is very sensible since NNs tend to generate points that cluster around a single prototype for each class \cite{snell2017prototypical}. 
Then, we fit a GP to each internal node which makes a binary decision based on the data associated with that node. We denote the GP associated with node $v_i$ by $f_{v_i} \sim\mathcal{GP}(m_{v_i}, k_{v_i})$, and all of the GPs in the tree with $\bld{\mathcal{F}}$. The likelihood of a data point having the class $c$ is given by the unique path $P^c$ from the root to the leaf node corresponding to that class:
\begin{align} \label{eq:tree_likelihood}
    & p(y=c | \bld{\mathcal{F}}) = \prod_{v_i \in P^c} \sigma(f_{v_i})^{y_{v_i}} (1 - \sigma(f_{v_i}))^{1 - y_{v_i}},
\end{align}
where $y_{v_i}=1$ if the path goes  left at $v_i$ and zero otherwise.



\subsection{Inference at the Node Level} \label{sec:method_inference_node}
Since the likelihood in \eqnref{eq:tree_likelihood} factorizes over the nodes, we may look at the individual components separately. In the following we omit the subscript $v_i$ for clarity; however, all datum and quantities are those that belong to a specific node $v_i$. In general, we can perform inference on each tree node with either Gibbs sampling or variational inference (VI). For training on large datasets with deep kernel learning and inducing points, we found the variational approach to scale better, as the inducing points posterior depends on the entire dataset. However, when modeling the novel classes in incremental few-shot learning, the Gibbs sampling procedure is more suitable, as no new parameters are required.

For Gibbs sampling, we use the posterior probabilities introduced in \secref{sec:backgroud_pg}. At each node, we can use block-Gibbs sampling to sample $\bomega$ and $\bff$. Then we can obtain the augmented marginal and predictive distributions described next. The augmented marginal likelihood:
\begin{equation} \label{eq:node_marginal_likelihood}
    \begin{aligned}
    p(\by | \bomega, \bX) &= \int p(\by | \bff, \bomega, \bX)p(\bff)d\bff \\
    &\propto \normal(\bld{\Omega}^{-1}\bld{\kappa} | \bld{0},~\bld{K} + \bld{\Omega}^{-1}),
    \end{aligned}
\end{equation}
and the augmented predictive likelihood on a new data point $\bx_*$:
\begin{equation} \label{eq:node_predictive_likelihood}
    \begin{aligned}
    p(y_* | \bx_*, \bomega, \by, \bX) &= \int p(y_* | f_*)p(f_* | \bx_*, \bomega, \by, \bX)df_*,\\
    \end{aligned}
\end{equation}
where, 
\begin{equation} \label{eq:node_predictive_posterior}
    \begin{aligned}
    p(f_* | \bx_*, \bX, \by, \bomega) &= \normal({f_* |\mu_*, ~\Sigma_*}), \\
    \mu_* &= \bld{k_*}^T(\bld{\Omega}^{-1} + \bld{K})^{-1}\bld{\Omega}^{-1}\bld{\kappa},\\
    \Sigma_* &= k_{**} - \bld{k_*}^T(\bld{\Omega}^{-1} + \bld{K})^{-1}\bld{k_*}.
    \end{aligned}
\end{equation}

Where we assumed a zero mean prior. The integral in \eqnref{eq:node_predictive_likelihood} is intractable but can be computed numerically with 1D Gaussian-Hermite quadrature.

Alternatively to the Gibbs sampling, we may apply variational inference at each node. 
We define $\bar{\bX}$ as the learned pseudo-inputs and  $\bar{\by}$ as their associated class labels. $\bar{\bX}$ are defined at the tree level and are shared by all relevant nodes.
For each node we define the variational distributions $q(\bomega) = PG(\bld{1}, \bld{c})$ and $q(\bar{\bld{f}}) = \normal(\bar{\bld{f}} | \bld{\tilde{\mu}}, \bld{\tilde{\Sigma}})$, where $\bld{c}, \bld{\tilde{\mu}}, \bld{\tilde{\Sigma}}$ are learnable parameters. The variational lower bound to the log marginal likelihood is:
\begin{equation} \label{eq:variational_lb}
    \begin{aligned}
    \mathcal{C}(\bld{c}, \bld{\tilde{\mu}}, \bld{\tilde{\Sigma}}) = \e_{p(\bff | \bar{\bld{f}})q(\bar{\bld{f}})q(\bomega)}[log~p(\bld{y}|\bomega, \bff)] - \\
    KL(q(\bar{\bld{f}}, \bomega) || p(\bar{\bld{f}}, \bomega)),
    \end{aligned}
\end{equation}
which has a closed-form expression.
We can then learn the model parameters and the variational parameters by maximizing the variational lower bound in \eqnref{eq:variational_lb}. 
The explicit form of \eqnref{eq:variational_lb} and the update rules of the variational parameters were adapted from \cite{WenzelGDKO19} and are presented in supplementary \secnref{sec:vi_bound_and_update}. 

To make predictions we can plug in the approximate posterior in the predictive posterior calculation:
\begin{equation} \label{eq:approximate_predictive}
    \begin{aligned}
    p(f_* | \bx_*, \bar{\bX}, \bar{\by}) &\approx \int p(f_* | \bar{\bld{f}}, \bx_*)q(\bar{\bld{f}})d\bar{\bld{f}} \\
    &= \normal(f_* | \mu_*,~\Sigma_*), \\
    \mu_* &= \bld{k}_{m*}^T \bld{K}_{mm}^{-1}\bld{\tilde{\mu}},\\
    \Sigma_* &= k_{**} - \bld{k_}{m*}^T (\bld{\tilde{\Sigma}}\bld{K}_{mm}^{-1} - \bld{I})\bld{k_}{m*},
    \end{aligned}
\end{equation}
where $\bld{k_}{m*}$ denotes the kernel vector between the inducing points and the test point, and $k_{**}$ denotes the kernel value at the test point.
Similarly to the Gibbs sampling case, we can obtain $p(y_* | x_*, \bar{\by}, \bar{\bX})$ by taking an integral over $f_*$ and compute it numerically with 1D Gaussian-Hermite quadrature.

\subsection{Full Learning of the Tree}\label{sec:learning_full}
We discussed how to learn and perform inference on a single node, we will now describe how to train the full tree model. For the full tree we need to learn the joint pseudo-inputs and the per-node variational parameters. Optimizing the full tree splits into the separate marginal likelihood of all examples and all the nodes on the path from the root to the leaf nodes:
\begin{equation}\label{eq:tree_objective}
    \begin{aligned}
        \mathcal{L} &= \sum_{j=1}^{n} \log~p(y_j | \bx_j, \bar{\by}, \bar{\bX})\\
        &= \sum_{j=1}^{n} \sum_{v_i \in P^{y_j}} \log~p(y_{v_i} | \bx_j, \bar{\by}, \bar{\bX}).
    \end{aligned}
\end{equation}
Since we cannot directly optimize this loss, we optimize the lower bound of it given in \eqnref{eq:variational_lb}. We summarize the learning algorithm of GP-Tree with VI in supplementary \secnref{sec:learning_alogrithm}. To apply predictions in the original multi-class problem, the prediction for a new data point $x_*$ is given as the product of predictive distributions from the root node to leaf node corresponding to each class.

Our method can be easily integrated with DKL. We simply superimpose the tree-based GP on an embedding layer of a neural network and learn the network parameters ${\theta}$ as well. In this case, we may define the inducing inputs in the input space or the feature space. We found that setting them in the feature space yields better performance, and requires less memory. Their location was initialized at the beginning of training using k-means++ applied on the embedding of data samples for each class separately. Also, we empirically found it beneficial to start the training process with a few epochs of standard training using the cross-entropy loss before building the GP tree and transitioning to learning with it.

\subsection{GP-Tree for Few-Shot Class-Incremental Learning}
\label{sec:method_incremental_learning}
\begin{figure}
    \centering
    \includegraphics[width=1.\linewidth]{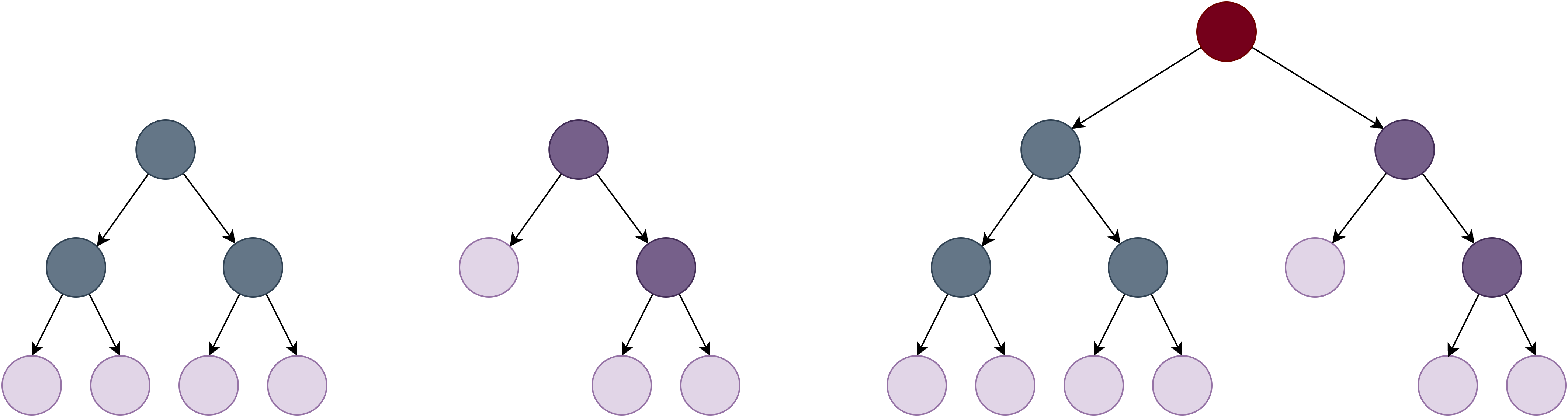}
    \text{
        \hspace{1pt} (a) Base 
        \hspace{28pt} (b) Novel
        \hspace{55pt}(c) Shared \hspace{50pt}
        }
    \caption{Tree expansion for novel classes: (a) A tree that was learned on the base classes; (b) On a novel session, we first build a tree from the novel classes representations; (c) Next we connect the trees using a shared root node.}
    \label{fig:tree_novel_classes}
\end{figure}
In class-incremental learning, we are given a sequence of labeled datasets $D_1, D_2, ..., D_T$, each sampled from a disjoint set of classes $C_1,..., C_T$. At each timestamp $t$, the model has access to dataset $D_t$ and the previous model, and it is tasked with classifying all the previously seen classes $\cup_{i=1}^t C_i$. In few-shot incremental learning \cite{RenLFZ19,tao2020fscil}, the base classes dataset, $D_1$, has a large number of samples, but all of the following datasets ($D_2$ onwards) have a limited amount of labeled data, e.g., 5 samples per class. Thus, after the model learned the previous classes it is tasked with learning new classes from few examples and without impairing the classification of previously seen classes whose data isn't available at that time, also known as catastrophic forgetting \cite{mccloskey1989catastrophic}.
Gaussian processes are naturally well-suited for this challenge. Bayesian models generalize well from few examples \cite{snell2020bayesian} and the inducing points can be used as a compact representation of the base classes, allowing us to avoid catastrophic forgetting. As Gaussian processes are non-parametric learners, we can classify new classes without fitting any new variables or tuning the NN parameters, which may also mitigate the catastrophic forgetting problem.

At the initial phase of learning the base classes, we employ GP-Tree (with DKL) to learn this dataset using VI as described earlier. We can view the inducing inputs, which are learned per class, as 'exemplars' of the base classes, a commonly used practice in incremental learning studies \cite{he2018exemplar}. We then use them in future learning sessions as described next. At the end of this stage, we freeze the NN backbone to avoid any parameter learning.

At each new session, we are given data from novel classes. We use the (fixed) NN backbone to obtain representations for the new samples that will be used for modeling the novel classes with GP-Tree. In this case, unlike the learning of base classes, the datasets are small. Therefore, we do not need to use inducing points. We just need to restructure the tree to account for the new classes and fit a GP to each new node. There are several alternatives for restructuring the tree, here we propose the following. We retain the original tree that was learned on the base classes intact, and at each novel session, we build a sub-tree from the embedding of samples in all (novel) sessions until the current one, namely $D_2,..., D_t$. We refer to these examples as novel examples hereafter. We then connect the sub-tree of novel examples with the sub-tree of the base classes with a shared root node. We define a GP classifier for each node in the sub-tree of novel examples and the root node. 
The GPs for the nodes in the sub-tree of the base classes are left untouched. Fitting the GPs in the sub-tree of novel examples is easily done with the representation of all examples at hand. For the root node, since we no longer have the base classes examples, we may use the inducing inputs of the base classes along with the embeddings of the novel examples. See illustration in \figrref{fig:tree_novel_classes}. We use Gibbs sampling on all new nodes to avoid any parameter tuning after the initial training on the base classes. Supplementary \secnref{sec_app:sensitivity} presents other alternatives for constructing the tree at novel sessions.

It is important to note that while we do save the inducing inputs and the network representations of samples from novel sessions for future sessions, this is in line with other incremental learning studies that save a few samples per class \cite{douillard2020podnet}. Furthermore, we only save the embeddings and not the original images. Therefore, the memory costs are low and in practice are negligible compared to storing the trained network weights. Also, we emphasize that the method described here is a natural way to extend our classification model to new classes. It was not tailored for the class-incremental few-shot scenario specifically. Despite that, we show strong results compared to models designed specifically for this task, especially on later learning sessions (see \secnref{sec:CIL}). Finally, GP-Tree can be easily extended to standard incremental learning setups. One immediate option is to build a tree and learn inducing inputs only for the novel classes seen at each new session similarly to the learning applied for base classes with VI. Then this sub-tree can be connected to the previous tree with a shared root node. This strategy can be further improved by taking only the inducing inputs from all classes seen thus far at the end of each novel session to rebuild the entire tree. Then we can use either the VI approach or, preferably, our Gibbs sampling procedure. In this study, we focus on the few-shot case and we leave this extension to future work.


\section{Related Work} \label{sec:related_work}
\textbf{Incremental learning.}
Incremental learning (IL) aims at learning new data without forgetting old data, what is known as 'catastrophic forgetting' \cite{mccloskey1989catastrophic}.  Recently, a large body of research was done in that direction, most of which is based on methods that use NNs only. Notable studies that use a similar procedure to ours are \citet{titsias2019functional} which regularize subsequent tasks with a set of inducing points stored from previous tasks, and \citet{gidaris2018dynamic} which also freeze the feature extractor after learning the base classes and infer a classification
weight vector for novel categories.
Methods in this field can be categorized according to three types: (i) \textit{Regularization based} approaches impose regularization methods on the network to maintain past knowledge~\cite{Goodfellow2014AnEI, kirkpatrick2017overcoming, lee2017overcoming, chaudhry2018riemannian, schwarz2018progress,RenLFZ19}. For example, \citet{kirkpatrick2017overcoming} limit the update of the parameters when encountering new data based on the fisher matrix  ; (ii) \textit{Architectural based} methods suggest network architectures that are resilient to the catastrophic forgetting issue and can accommodate new tasks~\cite{mallya2018piggyback, mallya2018packnet, yoon2018lifelong, serra2018overcoming, taitelbaum2019network, Liu2020MoreCL}. For example, \citet{rusu2016progressive} and following it \citet{yoon2018lifelong} expend the network with each new task. When applying parameters update the former freezes the previous network while the latter retrain part of it; (iii) \textit{Rehearsal based} aims at preventing catastrophic forgetting by storing and replaying information from previous episodes~\cite{rebuffi2017icarl, castro2018end, wu2018memory, hou2019learning,zhai2019lifelong, liu2020mnemonics, Liu2020MoreCL}. \citet{rebuffi2017icarl} introduced the class-incremental learning setup. They used exemplars to maintain information of past data and applied the nearest-mean-of-exemplars classification rule. In this paper, we follow the protocol suggested by \cite{tao2020fscil} for few-shot class-incremental learning.

\setlength{\tabcolsep}{4pt}
\begin{table*}[!t]
\centering
\caption{Test accuracy on CUB-200-2011. Average over 10 runs ($\pm$ SEM). In bold: statistically significant best results  (p=0.05). }
\vskip 0.1in
\scalebox{0.88}{
    \begin{tabular}{l c c c c c c c c c c}\toprule
    \multirow{2}{*}{Method}
    & \multicolumn{9}{c}{Number of Classes} \\\cmidrule{2-10}
    &4 &6 &8 &10 &20 &30 &40 &50 &60 \\\midrule
    SBM-GP & 96.98$\pm$0.5 & 95.55$\pm$0.9 &92.11$\pm$0.8 &91.04$\pm$0.7 &83.59$\pm$0.5 &77.63$\pm$0.9 &67.03$\pm$0.8 &64.20$\pm$1.0 &60.69$\pm$1.0\\
    OVE &97.33$\pm$0.5 &96.60$\pm$0.9 &94.70$\pm$0.8 & 93.05$\pm$0.8 & -- & -- & -- & -- & --\\
    LSM &97.30$\pm$0.9 &96.95$\pm$0.8 &95.16$\pm$0.8 &93.09$\pm$0.9 &89.23$\pm$1.3 &84.44$\pm$0.6 &72.90$\pm$1.2 &69.82$\pm$0.9 &65.53$\pm$0.9 \\
    \midrule
    GP-Tree Rnd. (ours) &  97.80$\pm$0.8& 95.96$\pm$0.7 &93.49$\pm$0.9 &92.07$\pm$1.1 &85.32$\pm$1.6 & 77.45$\pm$1.0 & 68.97$\pm$0.8 & 62.77$\pm$1.2 & 58.49$\pm$0.8 \\
    GP-Tree (ours) & 97.93$\pm$0.7 & 97.15$\pm$0.6 &94.67$\pm$0.9 &93.57$\pm$0.9 &88.77$\pm$1.4 &83.87$\pm$0.8 &\textbf{75.66$\pm$0.8} &\textbf{72.87$\pm$0.8} &\textbf{69.92$\pm$0.6} \\ 
    \bottomrule
    \end{tabular}
}
\label{Tab:gibbs_results}
\end{table*}

\textbf{Gaussian process classification.} In GPs for classification tasks the likelihood is no longer Gaussian and therefore approximation-based approaches or Monte-Carlo-based approaches are needed. Some classic non-augmentation based methods include the Laplace approximation \cite{williams1998bayesian}, expectation-propagation \cite{minka2001family}, and the least square approach \cite{rifkin2004defense}.
We refer the readers to \cite{gp_book} for a more thorough review. Recently some approaches emerged that are based on the \PG augmentation \cite{polya_gamma}. \citet{linderman2015dependent} proposed to use the \PG in a stick-breaking process to reparameterize a multinomial distribution as a product of binomial distributions. \citet{WenzelGDKO19} proposed to use \PG augmentation with variational inference for binary classification tasks. \citet{galy2020multi} proposed to use the logistic softmax likelihood and derived a conditionally conjugate model based on three augmentation steps. We believe that the quality of the prediction and learning may degrade as a result of the cascade of approximations. \citet{snell2020bayesian} proposed to use the one-vs-each likelihood in a few shot setting. Their method does not scale well with the data and classes due to the inversion of a $CN \times CN$ matrix. We use a Gibbs sampling approach for learning novel classes at each node. Several studies suggested alternative, more efficient, posterior sampling techniques when the classes are imbalanced or when using sub-samples of the data \cite{nemeth2020stochastic, sachs2020posterior, sen2020efficient}. Using such techniques to improve the standard Gibbs sampler is out of the current study scope.

\textbf{Scalable GPs.} In recent years some attempts were made to make GP classification more scalable. Inducing points \cite{silverman1985some, quinonero2005unifying, sneldon_Gharamani_IP} are a popular method to handle large datasets.
\citet{hoffman2013stochastic} developed a stochastic optimization process for VI. \citet{hensman2015scalable} introduced a method for GPC within a variational inducing point framework. \citet{izmailov2018scalable} used tensor train decomposition for variational parameters that allowed increasing dramatically the number of inducing points. \citet{wilson2016stochastic} proposed to learn multiple GPs on an embedding space and combine them linearly before a Softmax function. Extending this method to the incremental learning setting is not immediate as there are learnable parameters for combining the classes. \citet{bradshaw2017adversarial} used a GP with the Robust-Max likelihood for robustness against adversarial examples. This method doesn't scale well with the number of classes as we will show in \secnref{sec:exp_vi}.  

\textbf{Hierarchical models.} Hierarchical classifiers are a popular design choice. \citet{morin2005hierarchical}, and following it \citet{mnih2008scalable} proposed a tree-based classifier for language modeling. Their method applies to situations where all words are known in advance, while we need the ability to dynamically adapt the tree with new classes. \citet{damianou2013deep} stacked multiple GPs to create a hierarchy of GPs. \citet{nguyen2019scalable} proposed a hierarchical model using a mixture of GPs to learn global and local information. However, it is not clear how to extend this method to incremental learning challenges. 

Hierarchical stick break process was used in several contexts as well. \citet{adams2010tree} proposed a tree-based stick break for clustering as an alternative to the standard sequential stick break. \citet{nassar2019tree} proposed to use a tree-structure stick break for linear dynamical systems. Both methods did not include any GP components and do not have the flexibility of our model to apply inference with either VI or Gibbs sampling.

\section{Experiments} \label{sec:experiments}
In our experiments, we first examine several aspects of our method compared to previous common GPC methods (Sections \ref{sec:exp_gibbs} \& \ref{sec:exp_vi}). Then we evaluate GP-Tree on the setting of class incremental few-shot learning (\secnref{sec:CIL}). 
In the supplementary material we provide full implementation details (\secnref{sec:experimental_setup}), ablation study, and further analysis (\secnref{sec:further_expriments}).

\subsection{Inference with Gibbs Sampling} \label{sec:exp_gibbs}
As described in \secnref{sec:method}, GP-Tree allows to do inference with Gibbs sampling. This method is preferable when modeling novel classes in incremental learning, as we do not want to optimize any parameters and the size of the datasets are small. We evaluated GP-Tree in this setup on the fine-grained classification dataset, Caltech-UCSD Birds (CUB) 200-2011 \cite{WelinderEtal2010}. The CUB dataset contains 11,788 images of bird species from 200 classes with approximately 30 examples per class in the training set. Here, we did not apply DKL, but rather we used the pre-trained features published by \cite{xian2018feature}. This allowed us to only compare the inference part of our model. 

We compared GP-Tree against the following baselines that also used the \pg augmentation to get a conditionally conjugate likelihood: \textbf{(1) Stick Break Multinomial GP (SBM-GP)} \cite{linderman2015dependent}: that used the stick-breaking process to convert a multinomial likelihood to a product of binomial likelihoods; \textbf{(2) Logistic-Softmax (LSM)} \cite{galy2020multi} a recent method for GPC based on the logistic-softmax likelihood; and \textbf{(3) One-vs-Each (OVE)} \cite{snell2020bayesian} a method for GPC proposed recently for few-shot learning. Because this method requires the inversion of a $CN \times CN$ matrix, we were able to run it with only a few classes. 

\tblref{Tab:gibbs_results} compares GP-Tree against the baseline methods at increasing number of classes starting from $4$ to $60$ (out of $200$). The results are the average test-set classification accuracy along with the standard error of the mean (SEM) over ten seeds which included randomization in the class selection. 
When the number of classes is small ($\leq$ 30) GP-Tree, LSM, and OVE are comparable. However, as the number of classes increases GP-Tree performs better. \tblref{Tab:gibbs_results} also shows a variant of GP-Tree in which a balanced tree is built based on a random split of the classes (\textbf{GP-Tree Rnd}). This variant performs similarly to the SBM-GP baseline, indicating the importance of the class split algorithm in GP-Tree. 
To gain a better understanding of that we depict in \figrref{fig:gp_tree} the tree generated by GP-Tree on the CIFAR-10 dataset compared to a (possible) tree that corresponds to the SBM-GP model. The figure shows that: (i) GP-Tree generates a more balanced tree; and (ii) GP-Tree generates a tree that is ordered by the semantic meaning. For example, motorized vehicles are on the right subtree of the root node while animals are on the left subtree.

Finally, empirically we noticed that the number of steps in the Gibbs sampling procedure has a minor effect on the model performance. Therefore, we believe that it indicates that the chains converge quickly. Supplementary \secnref{sec:gp_tree_inf_appendix} further shows improved results for GP-Tree with more parallel Gibbs chains. It also compares the VI approach with the Gibbs sampling procedure, showing a large performance gap in favor of the latter. This result strengthens our choice for using the Gibbs procedure during novel sessions when using GP-Tree for incremental learning.
\subsection{GPC with DKL} \label{sec:exp_vi}
For evaluating GP-Tree with DKL we used the CIFAR-10 and CIFAR-100 datasets \cite{krizhevsky2009learning}. We compared GP-Tree with the following popular baselines that applied GPC with DKL: \textbf{(1) Stochastic Variational Deep Kernel Learning (SV-DKL)} \cite{wilson2016stochastic} which learned multiple GPs, each on a different subset of the embedding space, and combined them with the Softmax function; and \textbf{(2) GPDNN} \cite{bradshaw2017adversarial} which used the Robust-Max likelihood \cite{hernandez2011robust}. We used ResNet-18 \cite{he2016deep} as the backbone NN with an embedding layer of size 1024 and trained the models for 200 epochs. 

\tblref{Tab:GPs_with_DKL} shows the average accuracy across 3 seeds for both datasets. From the table, both GP-Tree and SV-DKL achieve high accuracy; however, GP-Tree prevails on both datasets. We found that GPDNN was extremely sensitive to the learning rate choice and the hyper-parameter controlling the probability of labeling error and we could not get reasonable results for it on the CIFAR-100 dataset.

\begin{table}[!t]\centering
\caption{Test accuracy ($\pm$ SEM) of GPs with DKL}\label{tab: }
\vskip 0.1in
\scalebox{1.}{
    \begin{tabular}{l cc cc cc}\toprule
    Method && CIFAR-10 && CIFAR-100 \\\midrule
    GPDNN && 81.16 $\pm$ 0.1 && -- \\
    SV-DKL && 92.73 $\pm$ .05 && 70.61 $\pm$ 0.2 \\
    \midrule
    GP-Tree (ours) && \textbf{93.25 $\pm$ 0.1} && \textbf{72.11 $\pm$ .05} \\
    \bottomrule
    \end{tabular}
    }
\label{Tab:GPs_with_DKL}
\end{table}

We note that the SV-DKL method is less suited for incremental learning, as the model includes a linear mapping followed by a softmax to produce a distribution over the classes. Therefore, it needs to learn new parameters with each new session and risks catastrophic forgetting, unlike our approach where no new parameters are tuned.

\subsection{Few-Shot Class-Incremental Learning} \label{sec:CIL}
In this section, we evaluate GP-Tree on the challenging task of few-shot class-incremental learning (FSCIL). We compare with methods that were designed for this learning setup and show comparable, if not superior, results by simply applying GP-Tree. This indicates that Gaussian processes in general and  GT-Tree, in particular, are well suited and a natural approach to incremental few-shot learning.

\begin{table*}[!t]
\centering
\caption{Few-shot class-incremental learning results on CUB-200-2011. Test accuracy averaged over 10 runs.}
\vskip 0.1in
\scalebox{.9}{
\begin{tabular}{l cc cc cc cc cc cc cc cc cc cc c}
\toprule
\multirow{2}{*}{Method}
&\multicolumn{21}{c}{Sessions} \\\cmidrule{2-22}
&1& &2& &3& &4& &5& &6& &7& &8& &9& &10& &11 \\\midrule
iCaRL & 68.68 && 52.65 &&  48.61 && 44.16 && 36.62 && 29.52 && 27.83 && 26.26 && 24.01 && 23.89 && 21.16 \\
EEIL & 68.68 && 53.63 && 47.91 && 44.20 && 36.30 && 27.46 && 25.93 && 24.70 && 23.95 && 24.13 && 22.11\\
NCM & 68.68 && 57.12 && 44.21 && 28.78 && 26.71 && 25.66 && 24.62 && 21.52 && 20.12 && 20.06 && 19.87\\
TOPIC & 68.68 && 62.49 && 54.81 && 49.99 && 45.25 && 41.40 && 38.35 && 35.36 && 32.22 && 28.31 && 26.28\\
\midrule 
SDC &  64.10 && 60.58 && 57.00 && 53.57 && 52.09 && 49.87 && 48.20 && 46.38 && 44.04 && 43.81 && 42.39\\
PODNet & \textbf{75.93 } && \textbf{70.29} && \textbf{64.50} && 49.00 && 45.90 && 43.00 && 41.33 && 40.56 && 40.09 && 40.59 && 39.30\\
\midrule
GP-Tree (ours) & 73.73 && 68.24 && 64.22 && \textbf{59.61} && \textbf{56.39} && \textbf{53.40} && \textbf{51.14} && \textbf{49.32} && \textbf{47.03} && \textbf{45.86} && \textbf{44.48} \\

\bottomrule
\end{tabular}
}
\label{Tab:cub}
\end{table*}

We follow the benchmarks proposed in \cite{tao2020fscil}, using the CUB 200-2011 dataset and mini-Imagenet, a 100-class subset of the Imagenet~\cite{deng2009imagenet} dataset used in few-shot studies~\cite{vinyals2016matching, finn2017model}. We adopt the 10-way 5-shot setting for CUB, choosing 100 base classes, and splitting the remaining 100 classes into ten incremental sessions. For mini-ImageNet, we follow the 5-way 5-shot, with 60 base classes, for a total of nine sessions.

Since the data splits made public by~\cite{tao2020fscil} did not include a validation set, we pre-allocate a small portion of the base classes dataset for hyper-parameter tuning of GP-Tree, SDC \cite{yu2020semantic}, and PODNet \cite{douillard2020podnet} on both datasets. 
\begin{figure}[!t]
    \centering
    \includegraphics[width=.85\linewidth, clip]{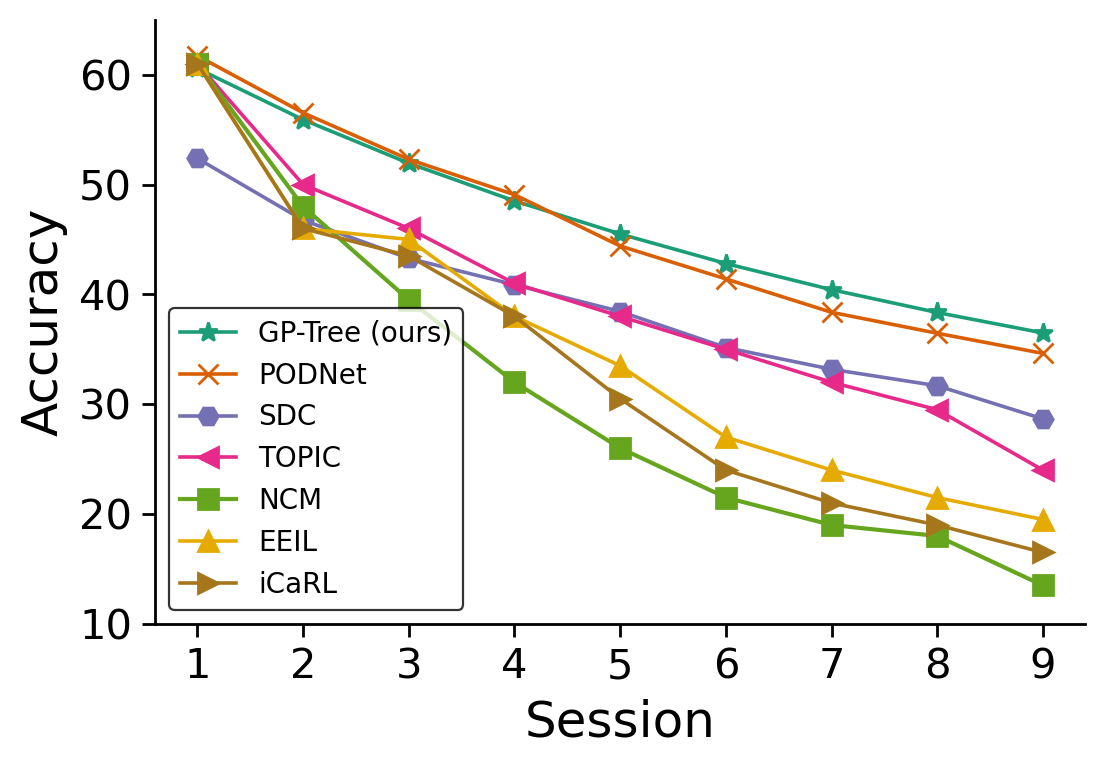}
    \caption{Few-shot class-incremental learning results on mini-ImageNet. Test set accuracy averaged over 10 runs. }
    \label{fig:mini_imagenet_acc}
\end{figure}

We compare GP-Tree with recent and leading FSCIL methods. The results of the following methods were taken from \cite{tao2020fscil}: 
\textbf{(1) iCaRL}~\cite{rebuffi2017icarl} that used exemplars of past data and applied the nearest-mean-of-exemplars classification; 
\textbf{(2) EEIL}~\cite{castro2018end} that used a distillation loss for old classes combined with a classification loss on all classes; 
\textbf{(3) NCM}~\cite{hou2019learning} which combined a classification loss, a distillation loss over normalized embedding layer, and a margin ranking loss; \textbf{(4) TOPIC}~\cite{tao2020fscil} that optimized a neural gas network with a classification loss, an anchor loss for less forgetting stabilization and a min-max loss to reduce overfitting.
We also compare with two additional baselines:
\textbf{(5) SDC}~\cite{yu2020semantic} that combined several losses to learn embedding representation and introduced a drift compensation to update previously computed prototypes; and \textbf{(6) PODNet}~\cite{douillard2020podnet} that used a spatial-based distillation-loss and a representation consisted of multiple proxy vectors per class.

The results for CUB are presented in \tblref{Tab:cub} and for mini-ImageNet in \figrref{fig:mini_imagenet_acc}. On both datasets, we found the PyTorch implementation of ResNet-18 to be consistent with the results seen in \cite{tao2020fscil}. We note that the results on mini-ImageNet could be improved by adapting the NN architecture for smaller images, but we kept the standard ResNet-18 for comparability with \cite{tao2020fscil}. 

The comparison shows that while PODNet outperforms all other methods during the first sessions, our GP-Tree achieves the best accuracy in the remaining sessions (4-11 in CUB and 5-9 in mini-ImageNet), where the challenges of avoiding catastrophic forgetting and learning from few-examples become more difficult. We note that for CUB experiments the average SEM was $0.4$ and for mini-ImageNet it was $0.2$. These results indicate that using GPs can indeed handle incremental learning challenges better than current procedures, but there is still room for improvement in how it handles the base classes. We also note that when examining the accuracy per session across all sessions at each time step, we noticed that GP-Tree showed a higher accuracy on the base classes compared to novel classes. This is an expected outcome since the number of novel examples is small and the feature extractor is kept fixed during novel sessions. Improving the way GP-Tree handles novel sessions can further boost its performance. In supplementary \secnref{sec_app:sensitivity} we perform sensitivity analysis on the kernel function choice and the number of representative samples stored per class. We show that non-linear kernel functions are preferred over a linear kernel function. In supplementary \secnref{sec_app:forgetting} we evaluate GP-Tree against baseline methods according to the average forgetting \cite{chaudhry2018riemannian}. We show that GP-Tree surpasses baseline methods on this aspect as well.

\section{Conclusion} \label{sec:conclusion}
In this work, we showed how common Gaussian process classification methods struggle when facing classification tasks with a large number of classes. We presented our method, GP-Tree, that scales with the number of classes and to large datasets. GP-Tree uses the \pg augmentation and allows great flexibility in posterior inference that can be done either with a variational inference approach or a Gibbs sampling procedure. We further showed how GP-Tree can be naturally and successfully combined with DKL. Finally, we demonstrated how GP-Tree can be adjusted to few-shot class-incremental learning challenges and showed how it achieves improved accuracy over current leading baseline methods. This indicates that Gaussian processes are a new and promising approach for this task.

\section*{Acknowledgements} 
This study was funded by a grant to GC from the Israel Science Foundation (ISF 737/2018), and by an equipment grant to GC and Bar-Ilan University from the Israel Science Foundation (ISF 2332/18). IA  was funded by a grant from the Israeli innovation authority, through the AVATAR consortium.

\clearpage
\bibliography{main}
\bibliographystyle{icml2021}
\clearpage

\twocolumn[
\icmltitle{Supplementary Material for GP-Tree: A Gaussian Process Classifier for Few-Shot Incremental Learning}]

\appendix
\section{Variational Bound \& Updates} \label{sec:vi_bound_and_update}
In \secnref{sec:method_inference_node} we presented the following variational lower bound for the log marginal likelihood at each node:
\begin{equation}
    \notag
    \begin{aligned}
    \mathcal{C}(\bld{c}, \bld{\tilde{\mu}}, \bld{\tilde{\Sigma}}) = \e_{p(\bff | \fbar)q(\fbar)q(\bomega)}[log~p(\bld{y}|\bomega, \bff)] - \\
    KL(q(\fbar, \bomega)~||~p(\fbar, \bomega)).
    \end{aligned}
\end{equation}
Here, we present the closed-form expression of it and the update rules for the variational parameters $\bld{\tilde{\mu}},\bld{ \tilde{\Sigma}}$ and $\bld{c}$. In the following constants are omitted for conciseness.

\subsection{Explicit Form for the Variational Bound}
We begin with the expectation term:
\begin{equation}
    \notag
    \begin{aligned}
    &\e_{p(\bff | \fbar)q(\fbar)q(\bomega)}[log~p(\bld{y}|\bomega, \bff)] \\ 
    &\propto  \e_{p(\bff | \fbar)q(\fbar)q(\bomega)}[(\bld{y} - \bld{1/2})^T\bff -\frac{1}{2}\bff^T\BigOmega\bff] \\
    &= \e_{q(\fbar)q(\bomega)}[(\bld{y} - \bld{1/2})^T \Knm\KmmInv\fbar - \frac{1}{2}Tr(\BigOmega \Qnn)\\ & \quad -\frac{1}{2}\fbar^T\KmmInv\Kmn\BigOmega\Knm\KmmInv\fbar]\\
    &= \frac{1}{2}\{2(\bld{y} - \bld{1/2})^T \Knm\KmmInv -Tr(\BigLambda\Qnn)\\ 
    &\quad -Tr(\KmmInv\Kmn\BigLambda\Knm\KmmInv\BigSigma)\\
    &\quad - \bld{\tilde{\mu}}^T\KmmInv\Kmn\BigLambda\Knm\KmmInv\bld{\tilde{\mu}}\},
    \end{aligned}
\end{equation}
where $\lambda_i = \e_{q(\omega_i)}[\omega_i] = \frac{1}{2c_i}tanh(\frac{c_i}{2})$, 
$\BigLambda = diag(\lambda_i)$. 

Now, we move to the KL divergence term. Due to independence between $p(\bld{\omega})$ and $p(\fbar)$, and the mean-field as a variational family assumption we have:
\begin{equation}
    \notag
    \begin{aligned}
    & KL(q(\fbar, \bomega)~||~p(\fbar, \bomega)) \\
    &= KL(q(\fbar) q(\bomega)~||~p(\fbar) p(\bomega)) \\
    &= KL(q(\fbar)~||~p(\fbar)) + KL(q(\bomega)~||~p(\bomega)).
    \end{aligned}
\end{equation}
The first KL term is between two Gaussian distributions and has the following closed-form expression:
\begin{equation}
    \notag
    \begin{aligned}
    & KL(q(\fbar)~||~p(\fbar))\\
    &= KL(\normal(\bld{\tilde{\mu}},~\BigSigma)~||~\normal(\bld{0},~\Kmm)) \propto\\ 
    & \frac{1}{2}\{Tr(\KmmInv\BigSigma) + \bld{\tilde{\mu}}^T\KmmInv\bld{\tilde{\mu}} - log~|\BigSigma| + log~|\Kmm|\}.
    \end{aligned}
\end{equation}

The second KL term is between two \pg (PG) distributions, each of a mutually independent random variable, and has a closed-form expression as well:
\begin{equation}
    \notag
    \begin{aligned}
    KL(q(\bomega) || p(\bomega))
    &= \sum_{i=1}^{n} KL(q(\omega_i)~||~p(\omega_i))\\
    &= \sum_{i=1}^{n} KL(PG(1,~c_i)~||~PG(1,~0))\\
    &= \sum_{i=1}^{n} log~cosh\frac{c_i}{2} - \frac{c_i}{4}tanh(\frac{c_i}{2}).
    \end{aligned}
\end{equation}

The variational lower bound is obtained by summing all these terms according to \eqnref{eq:variational_lb}.

\subsection{Variational Parameters Update}
The update rules for the variational parameters are given by taking the derivative of \eqnref{eq:variational_lb} w.r.t each of $\bld{c}, \bld{\tilde{\mu}}, \BigSigma$. At each iteration, based on a mini-batch of samples $\mathcal{B}$, we first update the parameters $\bld{c}^{\mathcal{B}} \subseteq {\bld{c}}$ corresponding to the samples seen in the batch using coordinate ascent scheme while holding $\bld{\tilde{\mu}}, \BigSigma$ fixed. Then, $\bld{\tilde{\mu}}, \BigSigma$ are updated according to a stochastic natural gradient ascent scheme.

The parameters $\bld{c}$ have a unique maximum which is given in a closed-form:
\begin{equation}
    \label{supp_eq:c_update}
    \begin{aligned}
    c_i = (Q_{ii} + \Kim\KmmInv\BigSigma\KmmInv\Kmi +\\ \bld{\tilde{\mu}}^T\KmmInv\Kmi\Kim\KmmInv\bld{\tilde{\mu}})^{\frac{1}{2}},
    \end{aligned}
\end{equation}
where the subscript $i$ denotes a specific row/column corresponding to the $i^{th}$ sample.

For the parameters $\bld{\tilde{\mu}}, \BigSigma$, the natural parameterization of the variational Gaussian distribution can be used: $\bld{\eta} = \BigSigma^{-1}\bld{\tilde{\mu}}$ and $\bld{H} = -\frac{1}{2}\BigSigma^{-1}$. The update at each batch then becomes:
\begin{equation}
    \label{supp_eq:natural_update}
    \begin{aligned}
    \tilde{\nabla}_{\bld{\eta}}\mathcal{C} &= \frac{n}{2|\mathcal{B}|}\KmmInv\Kmn^{\mathcal{B}}(\bld{y}^{\mathcal{B}} - \bld{1/2}) - \bld{\eta},\\
    \tilde{\nabla}_{\bld{H}}\mathcal{C} &= -\frac{1}{2}( \KmmInv + \frac{n}{2|\mathcal{B}|} \KmmInv\Kmn^{\mathcal{B}}\BigLambda^{\mathcal{B}}\Knm^{\mathcal{B}}\KmmInv) - \bld{H}.
    \end{aligned}
\end{equation}
Where we used the superscript $\mathcal{B}$ to denote only the rows/columns of samples in the batch. Note that the natural gradient updates maintain the positive-definiteness of $\BigSigma$.

\section{Learning Algorithm with VI} \label{sec:learning_alogrithm}
Algorithm \ref{algo:dkl_vi} summarizes GP-Tree learning with VI and DKL.

\begin{algorithm}[!t]
    \caption{GP-Tree Inference with VI} \label{algo:dkl_vi}
	\vspace{0.1cm}
	{\bf Input}: Data $\mathcal{\bld{D}} = (\bX, \by)$, $I_{1}$ number of training iterations with a NN, $I_{2}$ Number of training iterations with GP-Tree
	
	{\bf Init}: $g_{\theta}$ a NN parameterized by $\theta$
	
	{\bf For} $i = 1, ..., I_1$:\\
	\hspace*{5mm} - Sample a mini-batch of data from $\mathcal{\bld{D}}$\\
	\hspace*{5mm} - Learn $g_{\theta}$ with a classification loss\\
	{\bf End for}
    
    {\bf Build} GP-Tree $\mathcal{T}$ as described in \secnref{sec:hier_cls}
    
	{\bf Init}: GP hyper-parameters $\phi$, variational parameters $\bld{c}, \bld{\eta}, \bld{H}$, and inducing locations $\bar{\bld{X}}$ in the embedded space
	
	{\bf For} $i = 1, ..., I_2$:\\
        \hspace*{5mm} - Sample a mini-batch of data from $\mathcal{\bld{D}}$\\
        \hspace*{5mm} - Obtain embedding for $\bX$ with $g_\theta(\bX)$\\
        \hspace*{5mm} - Traverse the tree (e.g., via in-order traversal) \\
        \hspace*{5mm} {\bf For} each node in the path:\\
        \hspace*{10mm} - Update $\bld{c}$ according to \eqnref{supp_eq:c_update}\\
        \hspace*{10mm} - Update $\bld{\eta}, \bld{H}$ according to \eqnref{supp_eq:natural_update}\\
        \hspace*{5mm} {\bf End for}\\
        \hspace*{5mm} - Update $\theta$, $\phi$ and $\bar{\bld{X}}$ using \eqnref{eq:tree_objective} and by replacing the\\
        \hspace*{5mm} marginal likelihood terms with the variational lower\\ 
        \hspace*{5mm} bound $\mathcal{C}(\bld{c}, \bld{\tilde{\mu}}, \BigSigma)$ per node\\
    {\bf End for}\\
    {\bf Return} $g_{\theta}, \mathcal{T}, \bar{\bld{X}}$
\end{algorithm}

\section{Experimental Setup} \label{sec:experimental_setup}
This section provides further details about the experiments shown in \secnref{sec:experiments}.

\subsection{Inference with Gibbs Sampling - Sec. \ref{sec:exp_gibbs}}
\textbf{Data.} We used the pre-trained features extracted by \citet{xian2018feature} for the CUB 200-2011 dataset \cite{WelinderEtal2010}. The CUB 200-2011 dataset contains 200 classes of bird species in 11,788 images with approximately 30 examples per class in the training set. Here, since the training set size is limited, we used all 5994 training instances according to the official split and we split the predefined test set to 2897 samples for validation and 2897 for testing.

\textbf{Hyperparameter tuning.} For all baselines, in all experiments, we applied a grid search over the kernel type, either normalized linear kernel or normalized RBF kernel \cite{snell2020bayesian}. We consistently found that under this setting the linear kernel generated better results (this was not true in other settings). The output scale for the linear kernel was chosen based on a grid search in $\{1, 4, 9, 18\}$. We used $20$ Gibbs chains for the experiments with $\{4, 6, 8, 10, 20\}$ classes, $10$ Gibbs chains for the experiments with $30$ classes, and $1$ Gibbs chain for the experiments with $\{40, 50, 60\}$ classes. For the OVE baseline, we were able to use only $10$ chains for the experiments with $8$ classes and $3$ chains for the experiments with $10$ classes. These experiments were done on an NVIDIA V100 32GB GPU. We applied $1$ Gibbs sampling step before taking $\bld{\omega}$ for the predictive distribution calculations. In these experiments, we often found it useful to make predictions with a single sample at the expected value location instead of using the 1D Gaussian-Hermite quadrature.
\setlength{\tabcolsep}{3pt}
\begin{table*}[!t]
\centering
    \caption{Class-incremental few-shot learning on CUB-200-2011. Tree construction variants. Test accuracy averaged over 10 runs.}
    \vskip 0.1in
    \scalebox{.99}{
    \begin{tabular}{l cc cc cc cc cc cc cc cc cc cc c}
    \toprule
    \multirow{2}{*}{Method}
    &\multicolumn{21}{c}{Sessions} \\\cmidrule{2-22}
    &1& &2& &3& &4& &5& &6& &7& &8& &9& &10& &11 \\\midrule
    Session Tree & 73.73 && \textbf{68.24} && 64.07 && 59.42 && 56.12 && 52.80 && 50.73 && 48.89 && 46.34 && 45.01 && 43.33\\
    Rebuild Tree &73.73 && 67.71 && 63.50 && 59.03 && 55.73 && 52.73 && 50.49 && 48.85 && 46.17 && 44.84 && 43.20\\
    GP-Tree & 73.73 && \textbf{68.24} && \textbf{64.22} && \textbf{59.61} && \textbf{56.39} && \textbf{53.40} && \textbf{51.14} && \textbf{49.32} && \textbf{47.03} && \textbf{45.86} && \textbf{44.48}\\
    \bottomrule
    \end{tabular}
    }
\label{Tab:cub_construction_variants}
\end{table*}

\subsection{GPC with DKL - Sec. \ref{sec:exp_vi}}  
In all experiments we trained from scratch a ResNet-18 \cite{he2016deep} adjusted for CIFAR images size with a final embedding layer size of $1024$. The Batch size was set to $256$. We used SGD with a momentum of $0.9$ and a scheduler that decays the learning rate by a factor of $0.1$ at epochs $100$ and $150$. We allocated $10\%$ from the training set for validation using stratified sampling. We applied a grid search over the initial learning rate in $\{0.1, 0.01\}$ for all methods. We found that an initial learning rate of $0.01$ was preferred for our method. We used natural gradient descent with a learning rate of 0.05 for learning the variational parameters. In GPDNN experiments we also searched for an initial learning rate in $\{0.001, 0.0005\}$ and experimented with the Adam optimizer \cite{Kingma2014AdamAM}.
In all methods, we applied pre-training using a NN with a softmax layer after the last embedding layer and the cross-entropy loss. We searched over the number of pre-training epochs in $\{0, 20, 40, 60, 80\}$. For GP-Tree, $80$ epochs yielded the best results. We used $40$ inducing points per class in both GP-Tree and GPDNN experiments. For the SV-DKL baseline, we experimented with a grid size of $\{64, 128, 256\}$. In all experiments of all methods, we used the RBF kernel over L2 normalized input vectors. In CIFAR-100 experiments of the GPDNN baseline, we also applied an extensive grid search for the probability of labeling error without any success to achieve reasonable accuracy. In GP-Tree experiments, we found it beneficial to assign a weight to the loss term at each node that is inversely proportional to the amount of data used by that node for inference. It is achieved by dividing the loss at each node by the total number of training samples relevant for that node. 

\subsection{Few-Shot Class-Incremental Learning - Sec. \ref{sec:CIL}}
\label{sec:experimental_setup_FSCIL}
\textbf{Experimental protocol.} 
The experiments in this part largely followed the protocol suggested in \cite{tao2020fscil} for comparability. We adopted the 10-
way 5-shot setting for CUB, the first 100 classes were set as base classes, the remaining 100 classes were split into 10 incremental
sessions. For mini-ImageNet, we followed the 5-way 5-shot,
with 60 base classes, and 40 novel classes for a total of nine sessions. We used the official train/test split published by \citet{tao2020fscil}. We pre-allocated a small portion from the training set of the base classes for a validation set. From the CUB dataset, we took 2 samples per class. From the mini-ImageNet dataset, we allocated $5\%$ using stratified sampling. In CUB experiments we fine-tuned a pre-trained ResNet-18 on ImageNet while in mini-ImageNet experiments we trained it from scratch. The final embedding layer size was set to $512$. The mini-batch size at the first session was set to 128 in all experiments and, in later sessions, it included all available samples. We used SGD with a momentum of 0.9. 

\textbf{Hyperparameter tuning.}
In the experiments of GP-Tree, SDC \cite{yu2020semantic} and PODNet \cite{douillard2020podnet}, we applied a grid search over the initial learning rate at the first session in $\{0.1, 0.01, 0.001\}$. In CUB experiments it was set to $0.01$. In mini-ImageNet experiments, it was set to $0.1$. At the first session, we trained the models for $100$ epochs with a scheduler that decreased the learning rate by a factor of $0.1$ at epochs $40$ and $60$. At later sessions, for our method, there was nothing to set, as it doesn't require any learning. In SDC and PODNet experiments we trained for $100$ epochs and followed the training protocols suggested by each. For SDC, we applied a grid search over the embedding network in $\{LwF, MAS\}$, and $\gamma$, the hyper-parameter that controls the trade-off between the metric learning loss and the other losses, in $ \{5e-5, 5e-4, 5e-3, 5e-2, 5e-1, 1e4, 1e6\}$. For this baseline, we found it beneficial to start with a few epochs of training using a softmax layer and the cross-entropy loss and only afterward train with the triplet loss as advocated in the paper. When training with the triplet loss, we also needed to optimize the ratio of positive and negative examples at each batch to make it work.

\textbf{GP-Tree configurations.}
In GP-Tree experiments, at the initial session, we first applied a few epochs of training using a NN only with a softmax layer and the cross-entropy loss. Then, after $20$ epochs in CUB experiments, and $40$ epochs in mini-ImageNet experiments, we transitioned to learning with GP-Tree as described in \secnref{sec:method}. We used 5 inducing points per class and the RBF kernel on all GPs with an initial length-scale of $1$ and an initial output-scale of $4$. We learned the variational parameters with natural gradient descent and a learning rate of 0.05. Here, as well, the inputs to the kernel were normalized by their L2 norm. During training, we assigned a weight to the loss term at each node that is inversely proportional to the amount of data used by that node for inference. During novel sessions, we used the RBF kernel with a fixed length-scale of $1$ and a fixed output-scale of $8$. At the end of each few-shot session, we saved the $512$ dimensional representation of the samples for later sessions. 

\begin{figure}[!t]
\centering
    \begin{subfigure}[Gibbs vs VI]{
    \includegraphics[width=0.45\linewidth]{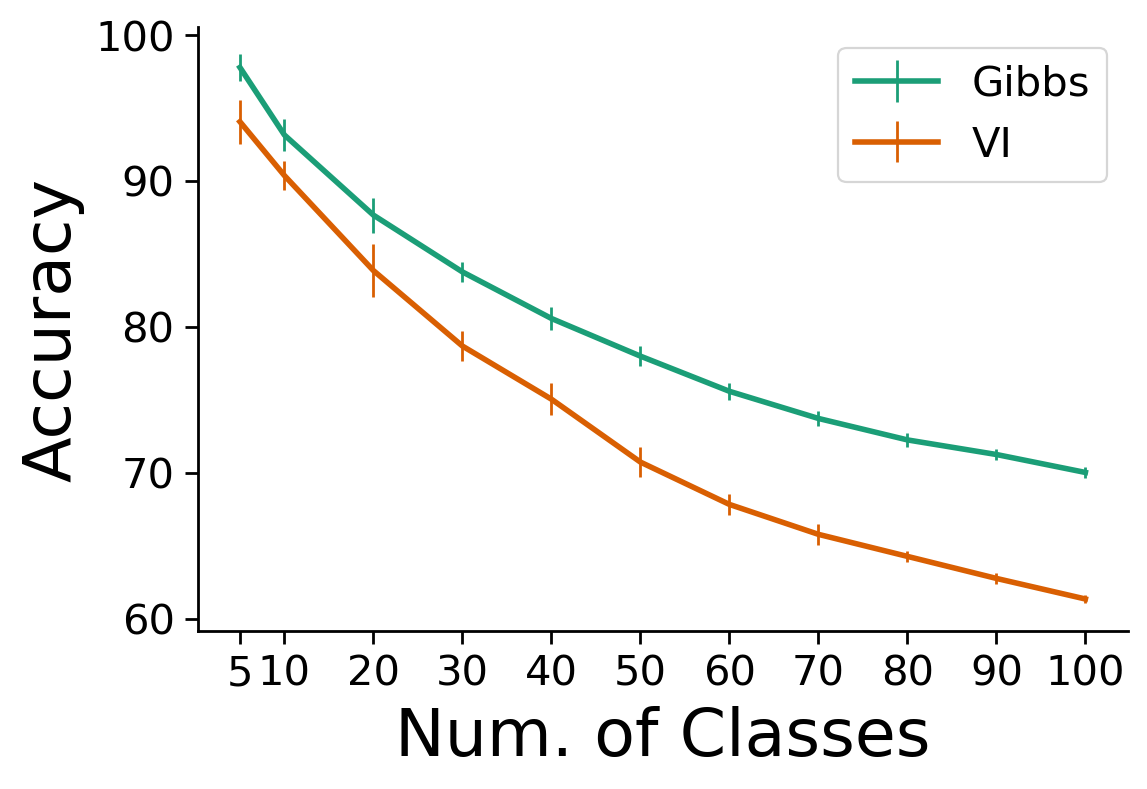}
    \label{fig:gibbs_vs_vi}
    }
    \end{subfigure}
    \begin{subfigure}[\# of Gibbs Chains]{
    \includegraphics[width=0.45\linewidth]{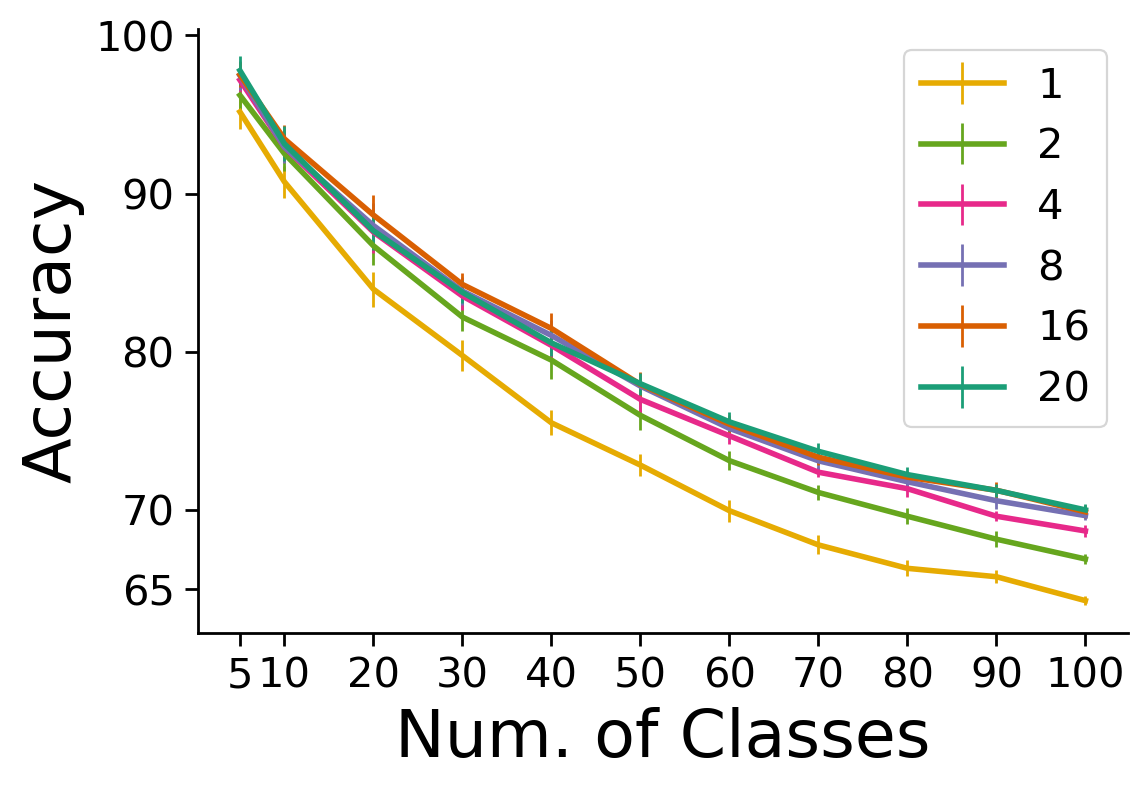}
    \label{fig:num_gibbs_chains}
    }
    \end{subfigure}
    \caption{Test accuracy when varying the number of classes. \textbf{Left} Gibbs sampling vs variational inference, \textbf{Right} varying the number of Gibbs chains. Results are the average over 10 runs ($\pm$ SEM) on pre-trained features of samples from the CUB 200-2011 dataset.}
\label{fig:gp_tree_inference}
\end{figure}

\section{Additional Experiments}\label{sec:further_expriments}
\subsection{GP-Tree Inference} \label{sec:gp_tree_inf_appendix}
The performance of GP-Tree depends on several factors. Here, we test (1) the effect of using VI against using a Gibbs sampler and, (2) the effect of the number of Gibbs chains. Both analyses were made by observing the test accuracy on the CUB-200-2011 dataset under the setup presented in \secnref{sec:exp_gibbs}. The results are shown in \figrref{fig:gp_tree_inference}. \figrref{fig:gibbs_vs_vi} shows a large performance gap in favor of the Gibbs sampling. This result is not surprising since the Gibbs sampler, asymptotically, samples from the correct posterior while in VI we use an approximate one. The figure also shows that the gap is amplified as the number of classes increases. This result is another justification for using Gibbs sampling when learning novel classes under the incremental learning setup. \figrref{fig:num_gibbs_chains} shows that as we increase the number of chains the accuracy increase as well; however, the difference is marginal when using four chains or more.

\subsection{Sensitivity Analysis for FSCIL} \label{sec_app:sensitivity}
When we adjusted GP-Tree to the few-shot class-incremental learning setting, we made several design choices. Here, we examine some of them. We will show that GP-Tree is fairly robust to these choices.

\textbf{Tree construction.} You may recall that under the FSCIL setup, after learning on the base classes, we retain the original tree intact, and at each novel session $t$, we build a sub-tree from samples' representations that appeared in sessions $D_2, ..., D_t$. We then connect this sub-tree to the base classes tree with a shared root node. In \tblref{Tab:cub_construction_variants} we present two alternatives for the tree construction used for inference during novel sessions ($t > 1$). (1) \textit{Session Tree}, instead of building a tree from samples in all sessions $D_2, ..., D_t$, in each session we build a tree from classes that appeared at the current session only. Then, we connect this sub-tree with the tree that is already built via a shared root; (2) \textit{Rebuild Tree}, building the entire tree at each new session and fitting the Gibbs sampling variant of our approach to each node. To account for base classes we use the inducing inputs.
\tblref{Tab:cub_construction_variants} shows that both alternatives yield good results; however, the approach chosen for GP-Tree is superior.

\begin{figure}[!t]
\centering
    \begin{subfigure}[\# of Rep. Samples]{
    \includegraphics[width=0.45\linewidth]{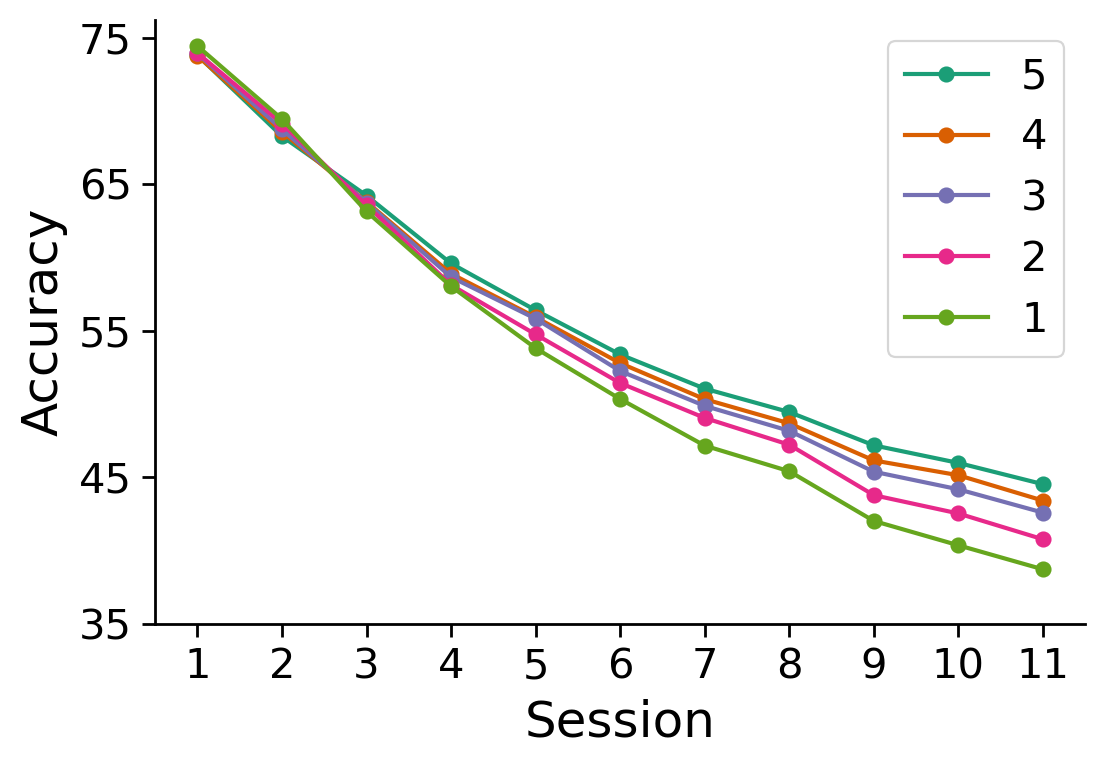}
    \label{fig:cub_num_inducing}
    }
    \end{subfigure}
    \begin{subfigure}[Kernel Choice]{
    \includegraphics[width=0.45\linewidth]{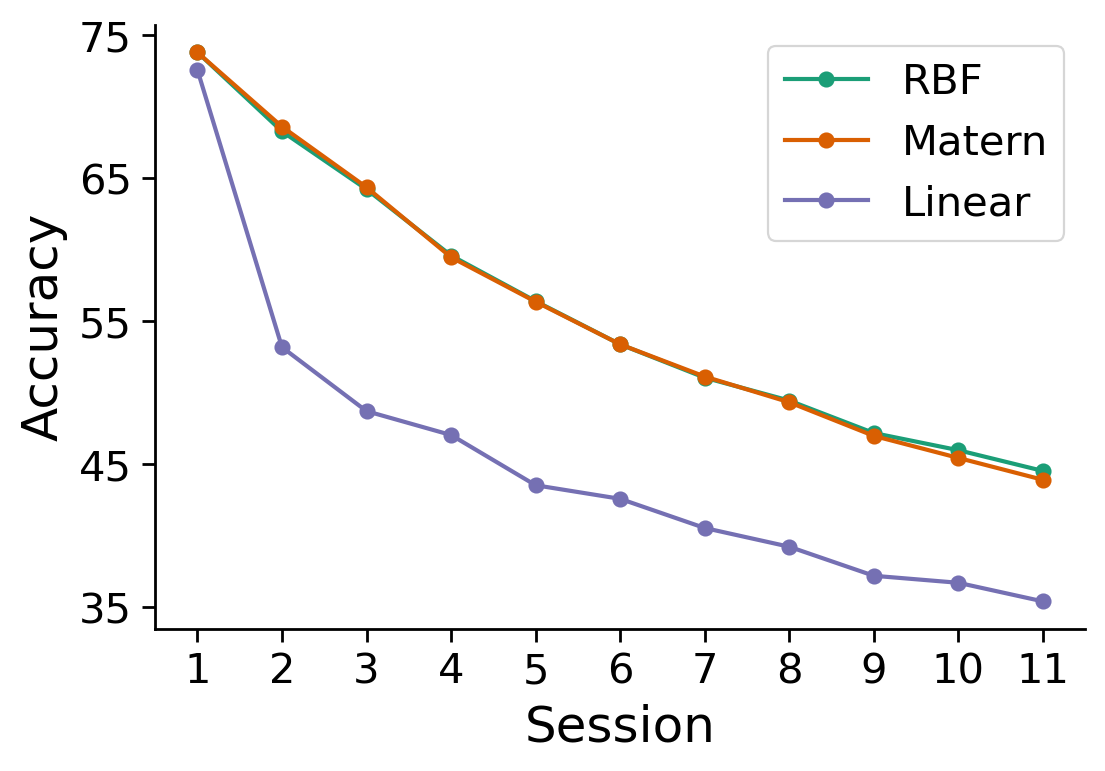}
    \label{fig:cub_kernel_choice}
    }
    \end{subfigure}
    \caption{Test accuracy for few-shot class-incremental learning as a function of the number of representative samples per class (left) and the kernel choice (right). Results are the average over 5 runs on the CUB 200-2011 dataset.}
\label{fig:cub_choices}
\end{figure}

\textbf{Kernel analysis.} The results presented in the main paper for few-shot class-incremental learning were with the RBF kernel and $5$ representative samples per class. Here we investigate both choices in \figrref{fig:cub_choices}. \figrref{fig:cub_num_inducing} compares between $1 - 5$ representative samples per class. The figure shows that all alternatives achieve high accuracy across all sessions; however, as expected, when using less representative samples there is a slight degradation in performance. The impact becomes more severe in later sessions. \figrref{fig:cub_kernel_choice} shows a comparison between the RBF, Matern 5/2 and Linear kernels. Similar to \cite{snell2020bayesian} we found gain in normalizing the inputs to the kernels by their L2 norm. Therefore, we applied this method in all settings and for all kernel choices. However, unlike in the few-shot learning case presented in \cite{snell2020bayesian}, in which the normalized linear kernel (also referred to as cosine kernel in that study) yielded the best results, in our case either the RBF kernel or the Matern kernel are preferred by a large margin, especially on novel sessions. We hypothesis that on the base classes the NN outputs a representation that is more linearly separable. Therefore, all kernels perform similarly on them. However, the representation of novel classes is more mixed in the embedding space. Therefore, a stronger kernel that generates non-linear decision boundaries is required.

\subsection{Forgetting Across Sessions} \label{sec_app:forgetting}
\begin{figure}[!t]
\centering
    \includegraphics[width=0.8\linewidth]{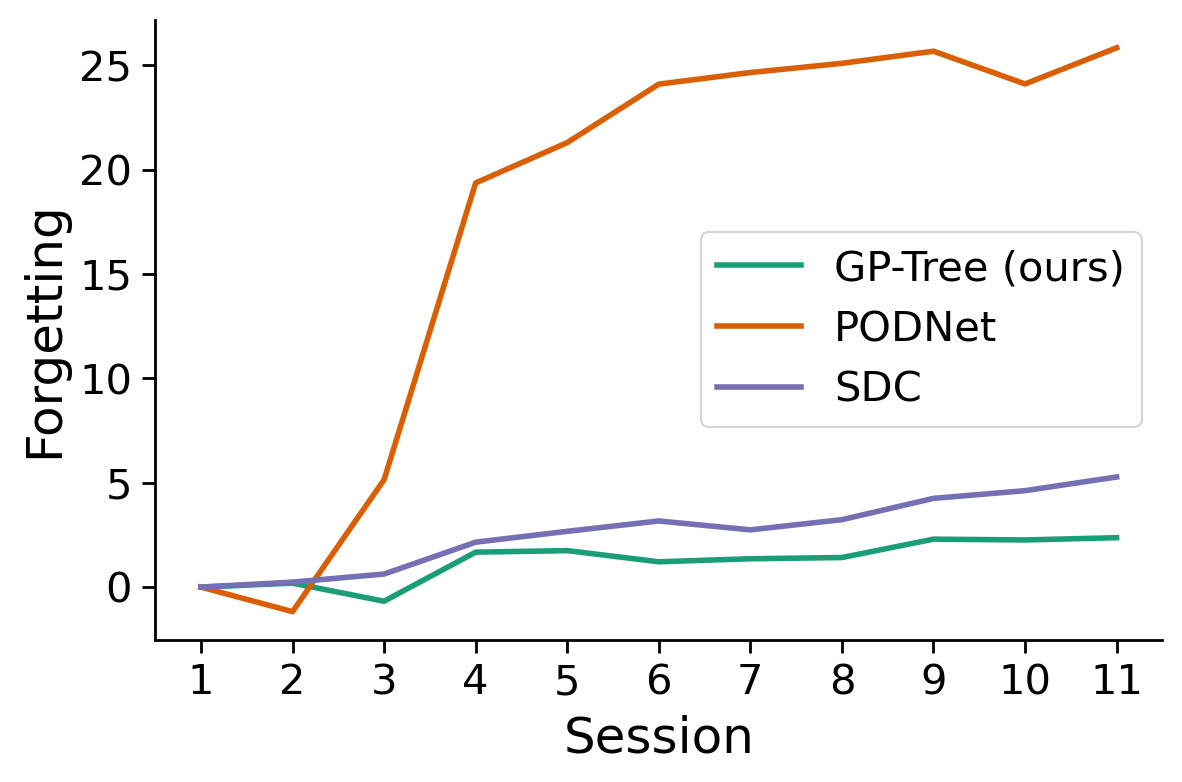}
    \caption{Average forgetting across sessions. Results are the average over 5 runs on the CUB 200-2011 dataset.}
    \label{fig:forgetting}
\end{figure}

In this part, we examine how GP-Tree performance varies across sessions through the \textit{average forgetting} \cite{chaudhry2018riemannian}. The average forgetting was designed to estimate the forgetting of prior tasks. Let $\alpha_j^k$ denote the accuracy of the learner on the $j^{th}$ task at session $k > j$. The forgetting of the $j^{th}$ task is defined as $g_j^k = \max\limits_{l \in \{1, ..., k-1\}} \alpha_j^l - \alpha_j^k$. This quantity is measured for every task $j$ seen thus far at each new session. Then, to get an estimate of the average forgetting we may use: $\frac{1}{k-1}\sum_{j=1}^{k-1}g_j^k$. \figrref{fig:forgetting} shows the average forgetting across all classes and sessions on the CUB 200-2011 dataset for GP-Tree, SDC \cite{yu2020semantic}, and PODNet \cite{douillard2020podnet}. From the figure, we notice a minor forgetting for GP-Tree. When comparing the performance of the three methods across all sessions, we noticed that GP-Tree and SDC showed better performance on the base classes while PODNet was more balanced with a slight advantage to the novel classes.

\end{document}